\documentclass{article}
\usepackage{spconf,amsmath,epsfig}
\usepackage{multirow} 
\usepackage{amsfonts}
\usepackage{booktabs}
\usepackage{threeparttable}
\usepackage{bm}
\usepackage[marginal]{footmisc}
\usepackage[preprint]{spconfa4}

\usepackage{algorithm}
\usepackage{algpseudocode}
\usepackage{amsmath}
\begin{document}\sloppy

% Example definitions.
% --------------------
\def\x{{\bm x}}
\def\L{{\cal L}}

% Title.

% Title.
% ------
\title{Beyond without Forgetting: Multi-Task Learning for Classification with Disjoint Datasets}
%
% Address.
% ---------------

\name{Yan Hong, Li Niu, Jianfu Zhang, Liqing Zhang}

\address{}

\maketitle

\begin{abstract}
Multi-task Learning (MTL) for classification with disjoint datasets aims to explore MTL when one task only has one labeled dataset. In existing methods, for each task, the unlabeled datasets are not fully exploited to facilitate this task. Inspired by semi-supervised learning, we use unlabeled datasets with pseudo labels to facilitate each task. However, there are two major issues: 1) the pseudo labels are very noisy; 2) the unlabeled datasets and the labeled dataset for each task has considerable data distribution mismatch. To address these issues, we propose our MTL with Selective Augmentation (MTL-SA) method to select the training samples in unlabeled datasets with confident pseudo labels and close data distribution to the labeled dataset. Then, we use the selected training samples to add information and use the remaining training samples to preserve information. Extensive experiments on face-centric and human-centric applications demonstrate the effectiveness of our MTL-SA method.
\end{abstract}
\begin{keywords}multi-task learning, semi-supervised learning, pseudo label
\end{keywords}
\section{Introduction} \label{sec:intro}
Multi-task Learning (MTL) for classification targets at exploiting the shared information among multiple related tasks. Each classification task (\emph{e.g.}, gender classification) has a corresponding label set (\emph{e.g.}, male and female). In standard MTL, one dataset is usually associated with multiple label sets corresponding to multiple tasks. However, in the real world, one dataset is often associated with only one label set corresponding to one task, and thus multi-task learning requires multiple disjoint datasets.
%It is meaningful to train a multi-task network with datasets of which sample only has a label for one task, 
This learning scenario is called multi-task learning with disjoint datasets in our paper. 
%The difference between standard MTL and MTL with disjoint datasets is illustrated in Figure~\ref{fig:scenario}. 

%\begin{figure}
%\begin{center}
%\includegraphics[scale=0.25]{task_scenario.png}
%\end{center}
%\caption{The difference between standard multi-task learning and multi-task learning with disjoint datasets. Best viewed in color.}
%\label{fig:scenario} 
%\end{figure}

In MTL with disjoint datasets, the training strategies can be roughly categorized into joint training and alternating training. For ease of representation, in the remainder of this paper, we take two datasets corresponding to two tasks as an example, in which dataset A (\emph{resp.}, B) is used for task A (\emph{resp.}, B). In joint training, dataset A and dataset B are jointly used to train a multi-task network at the same time~\cite{ranjan2017all}. However, in this case, task A (\emph{resp.}, B) would be biased towards the labeled dataset A (\emph{resp.}, B) instead of benefiting from the existence of unlabeled dataset B (\emph{resp.}, A) \cite{zhang2018facial}. 

Instead of joint training, a naive alternating training strategy is as follows. Suppose dataset B is used in the current epoch, the training process for task B is supervised by dataset B with ground-truth labels, while task-specific parameters for task A remain frozen. However, the training process forgets knowledge learned from dataset A in the previous epoch when using dataset B. To preserve the knowledge, Learning without Forgetting (LwF)~\cite{li2018learning} proposed to supervise task A by dataset B with soft label vector (i.e., decision values of all categories) predicted by the model trained on dataset A in the previous epoch. We refer to this alternating training strategy as MTL-wF. However, the drawback of MTL-wF is that when using dataset B, the training process for task A only preserves the information of dataset A from the previous epoch without fully exploiting the information in dataset B.

In this paper, we build our method upon MTL-wF, but aim to fully exploit the unlabeled dataset to add information instead of only preserving information. Tagging unlabeled data with pseudo label to augment training data has been widely used in semi-supervised learning~\cite{chapelle2009semi,lee2013pseudo,berthelot2019mixmatch}. Therefore, we tend to augment each task using the unlabeled dataset with pseudo labels. By taking the epoch of using dataset B as an example, we can obtain the soft label vectors of dataset B for task A, which are predicted by the model trained on dataset A in the previous epoch. The soft label vector can be converted to one-hot pseudo label vector, similar to semi-supervised learning~\cite{chapelle2009semi,lee2013pseudo,berthelot2019mixmatch}. Then, we can use dataset B with pseudo labels as additional supervision information to augment task A. However, there exist two major issues. Firstly, predicted pseudo labels could be very noisy. When using the training samples in dataset B with very noisy pseudo labels to augment task A, the performance on task A could be adversely affected. Secondly, the data distributions of dataset A and dataset B may be considerably different. Following the terminology in domain adaptation~\cite{PatelGLC15}, dataset A and dataset B with different data distributions can be referred to as domain A and domain B respectively. When applying the model trained on domain B to the test samples in domain A, the performance could be significantly degraded due to the data distribution mismatch~\cite{TorralbaE11}.

We will discuss how to address the above two issues by taking the epoch of using dataset B as an example. To address the first issue, we tend to use confidence score and local density to select the training samples in dataset B with confident pseudo labels. To address the second issue, we cluster dataset B into different groups and select those groups with closer data distribution to dataset A based on data distribution difference. To simultaneously handle the above two issues, we tend to select the training samples in dataset B which have both confident pseudo labels and close data distribution to dataset A. For the selected samples, we use pseudo label vectors as their training label vectors to add information. For the unselected training samples, we use soft label vectors as their training label vectors to preserve information as in~\cite{li2018learning}. Instead of binary selection, we assign different weights for different training samples in dataset B and the assigned weights are used to interpolate pseudo label vectors and soft label vectors, leading to interpolated label vectors. By using the interpolated label vector as training label vector, each training sample in dataset B can augment task A to different degrees. Therefore, we name our method as  Multi-Task Learning with Selective Augmentation (MTL-SA). Comprehensive experiments on four face-centric datasets and two human-centric datasets demonstrate the superiority of our MTL-SA.

%Our major contributions are summarized as follows: 1) We propose our Multi-Task Learning with Selective Augmentation (MTL-SA) method for MTL with disjoint datasets to fully exploit the unlabeled datasets for each task; 2) Technically, we augment each task with the selected training samples from unlabeled datasets, which have confident pseudo labels and close data distribution to the labeled dataset; 3) Comprehensive experiments on four face-centric datasets and two human-centric datasets demonstrate the superiority of our proposed MTL-SA.

%%%%%%%%% BODY TEXT
\section{Related Work}
%In this section, we will review the related works on multi-task learning (MTL) with disjoint datasets, semi-supervised MTL, and label vector interpolation.
%-------------------------------------------------------------------------
\subsection{Multi-task Learning with Disjoint Datasets}
%Multi-task learning (MTL) has achieved great progress in many applications. 
%In standard MTL, one dataset is associated with multiple label sets corresponding to multiple tasks. Besides standard MTL, one dataset could be only associated with one label set corresponding to one task, in which case one task only has one labeled dataset. 
The training strategies for MTL with disjoint datasets can be roughly categorized into joint training and alternating training: 1) For joint training, the methods in~\cite{chen2016neural,hashimoto2016joint} proposed to treat all tasks equally and train the multi-task network with disjoint datasets; 2) For alternating training, the method in~\cite{Bilen2017Universal} proposed to use one dataset to supervise corresponding task in each epoch. The idea of \cite{li2018learning} is adopted in \cite{kim2018disjoint}, which aims to preserve information from the previous epoch. All of the above works do not fully exploit unlabeled datasets for each task while our method can select training samples from unlabeled datasets to augment each task.

%-------------------------------------------------------------------------
\subsection{Semi-supervised Multi-task Learning}
%A few works have attempted to combine semi-supervised learning and multi-task learning (MTL). 
One group of semi-supervised MTL methods~\cite{lu2015semi,chang2017semisupervised} exploit shared manifold information among multiple tasks. Another group of semi-supervised MTL methods~\cite{zhang2018facial,augenstein2018multi,yang2019deep,du2019doubly} aim to infer confident pseudo labels for unlabeled training samples. Our method is more related to the second group. Although the above methods consider how to infer confident pseudo labels, they do not consider the data distribution mismatch between labeled and unlabeled training samples. In contrast, our method considers both pseudo label noise and data distribution mismatch when using unlabeled training samples. 

\subsection{Label Vector Interpolation} 
The goal of label vector interpolation is incorporating different types of label information to smooth label vector or handle the label noise. To name a few, Szegedy \emph{et al.}~\cite{szegedy2016rethinking} proposed to interpolate the label vector and a constant vector with uniform values to smooth the label vector. Li \emph{et al.}~\cite{li2017learning} proposed to interpolate the noisy label vector and the label vector predicted by an auxiliary model trained on clean data to handle the label noise. %Similarly, Reed \emph{et al.}~\cite{reed2014training} interpolated the noisy label vector and the label vector predicted by its own model to deal with the label noise. 
However, they use the same interpolation coefficient for all training samples. Instead, our method assigns different interpolation coefficients to different training samples adaptively.

\section{Background}
%In this section, we provide the problem definition of MTL with disjoint datasets and describe the alternating training strategy for MTL with disjoint datasets.
\subsection {Problem Definition}
In MTL for classification with disjoint datasets, we assume that we have two datasets corresponding to two tasks. Images $\mathcal{D}_A=\{\mathbf{I}^a_1, ..., \mathbf{I}^a_{n_a}\}$ from dataset A are labeled with $\mathcal{Y}_A=\{\mathbf{y}^a_1, ..., \mathbf{y}^a_{n_a}\}$ with the label set corresponding to task A, while images $\mathcal{D}_B=\{\mathbf{I}^b_1, ..., \mathbf{I}^b_{n_b}\}$ from dataset B are annotated with $\mathcal{Y}_B=\{\mathbf{y}^b_1, ..., \mathbf{y}^b_{n_b}\}$ with the label set corresponding to task B. Our multi-task network  consists of convolutional layers with model parameter $\bm{\theta}^s$ shared by two tasks and task-specific layers with model parameter $\bm{\theta}^a$ (\emph{resp.}, ${\bm{\theta}}^b$) for task A (\emph{resp.}, B). Besides, we use $p_t^a(\cdot)$ (\emph{resp.}, $p_t^b(\cdot)$) to represent the label predictor based on model parameters $\{\bm{\theta}^{s},\bm{\theta}^{a}\}$ (\emph{resp.}, $\{\bm{\theta}^{s},\bm{\theta}^{b}\}$). Similarly, we use $f_t^a(\cdot)$ (\emph{resp.}, $f_t^b(\cdot)$) to represent the feature extractor based on model parameters $\{\bm{\theta}^{s},\bm{\theta}^{a}\}$ (\emph{resp.}, $\{\bm{\theta}^{s},\bm{\theta}^{b}\}$) with the last classification layer removed.
%In the training process, how to train such multi-task network effectively with disjoint datasets is worth studying. Since our method adopts alternating training strategy, 
Next, we will introduce the alternating training strategy with information preservation~\cite{li2018learning}.

\subsection {Multi-task Learning without Forgetting} \label{sec:alter_train}
\begin{figure}[t]
\begin{center}
\includegraphics[scale=0.3]{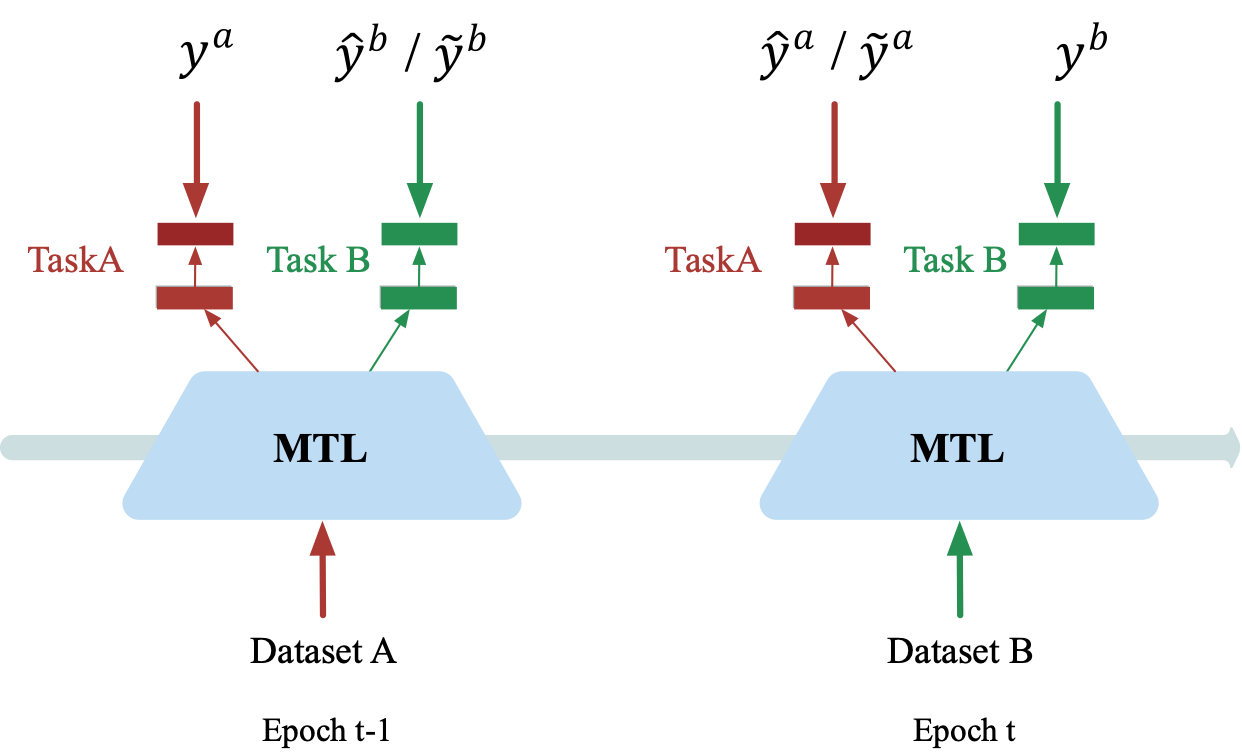}
\end{center}
\vspace{-15pt}
\caption{Alternating training strategy for multi-task learning with disjoint datasets. We only show two epochs here, in which dataset A is used in training epoch $t\!-\!1$ and dataset B is used in training epoch $t$.}
\label{fig:training_strategy}
\end{figure}

As discussed in Section 1, the idea of Learning without Forgetting (LwF)~\cite{li2018learning} could be incorporated into naive alternating training strategy, leading to Multi-Task Learning without Forgetting (MTL-wF). The process of MTL-wF is depicted in Figure~\ref{fig:training_strategy}. Specifically, training images $\mathcal{D}_A=\{\mathbf{I}^a_1, ..., \mathbf{I}^a_{n_a}\}$ from dataset A and images $\mathcal{D}_B=\{\mathbf{I}^b_1, ..., \mathbf{I}^b_{n_b}\}$ from dataset B are fed into multi-task network in an alternating fashion, in which $n_a$ (\emph{resp.}, $n_b$) is the number of training images in dataset A (\emph{resp.}, B). As shown in Figure~\ref{fig:training_strategy}, in epoch $t\!-\!1$, the network is trained with images $\mathcal{D}_A$ from dataset A. Each image $\bm{I}^a_i$ has ground-truth label vector $\bm{y}_i^a$ for task A, but does not have ground-truth label for task B. Thus, we use the decision values of $\bm{I}^a_i$ activated by label predictor $p_{t-2}^b(\cdot)$ (\emph{i.e.}, $\{\bm{\theta}^{s},\bm{\theta}^{b}\}$ from  epoch $t\!-\!2$) as the soft label vector $\bm{\tilde y}^{b}_i$ of  $\bm{I}^a_i$. 
Subsequently, images $\mathcal{D}_B$ from dataset B are used to train the network in epoch $t$, in which each image $\bm{I}^b_i$ has ground-truth label vector $\bm{y}_i^b$ for task B and soft label vector $\bm{\tilde y}^{a}_i$ activated by label predictor $p_{t-1}^a(\cdot)$ for task A. In alternating training, dataset B (\emph{resp.}, A) with soft label vector are used for task A (\emph{resp.}, B).
The reason of using soft label vector as supervision is that the task-specific layers will become less effective if the shared layers are updated while the task-specific layers remain unchanged, which is dubbed as forgetting effect~\cite{li2018learning}. Formally, with the soft label $\bm{\tilde y}_i^{b} = p_{t-2}^b(\bm{I}^a_i)$ of $\mathcal{D}_A$, the loss function in epoch $t\!-\!1$ can be written as

\vspace{-10pt}
\begin{equation}
\begin{aligned}
\min \limits_{\bm{\theta}^s,\bm{\theta}^a,\bm{\theta}^b} \sum_{i=1}^{n_a} L(\bm{y}_i^a,p_{t-1}^a(\bm{I}^a_i)) + L(\bm{\tilde y}_i^b,p_{t-1}^b(\bm{I}^a_i)),
\label{naive_loss_a} 
\end{aligned}
\end{equation}
where $L(\bm{y},\bm{p})$ is the cross-entropy loss calculated based on the input label vector $\bm{y}$ and the output decision values $\bm{p}$.

Similarly, in epoch $t$, with the soft label $\bm{\tilde y}^a_i=p_{t-1}^a(\bm{I}^b_i)$  of $\mathcal{D}_B$, the loss function  can be written as

\vspace{-10pt}
\begin{equation}
\begin{aligned}
\min \limits_{\bm{\theta}^s,\bm{\theta}^a,\bm{\theta}^b} \sum_{i=1}^{n_b} L(\bm{\tilde y}_i^{a},p_{t}^a(\bm{I}^b_i)) + L(\bm{y}_i^b,p_{t}^b(\bm{I}^b_i)). \label{naive_loss_b}
\end{aligned}
\end{equation}
As the number of training epochs increases, dataset A and dataset B are alternatingly fed into the multi-task network, and $\{\bm{\theta}^s,\bm{\theta}^a,\bm{\theta}^b\}$ can be updated continuously without forgetting effect. One similar work to MTL-wF is the method in \cite{kim2018disjoint}, but this method focuses on action classification and captioning instead of multi-classification tasks. 
%-------------------------------------------------------------------------
\section{Our Method}
In this section, we extend MTL-wF introduced in Section 3.2 to our Multi-Task Learning with Selective Augmentation (MTL-SA) method. Unlike MTL-wF which can only preserve information without forgetting effect, our method aims to utilize the unlabeled dataset to augment each task with extra information, like semi-supervised multi-task learning. Inspired by semi-supervised MTL methods using pseudo labels~\cite{zhang2018facial,augenstein2018multi}, we tend to generate pseudo labels for unlabeled dataset and use them as weak supervision information to augment each task. In particular, given a soft label vector, we can easily obtain the corresponding pseudo label vector by setting the entry with the highest decision value as $1$ and the remaining entries as $0$~\cite{lee2013pseudo}.  However, for each task, there exist two major issues when using the unlabeled dataset with pseudo labels to augment this task: pseudo label noise as well as the data distribution mismatch between the unlabeled dataset and the labeled dataset. So it may be ineffective to use all training samples in the unlabeled dataset with pseudo labels. To address the above two issues, we tend to select the training samples in the unlabeled dataset with confident pseudo labels and close data distribution to the labeled dataset. Then, we use pseudo label vectors for the selected training samples to add information while using soft label vectors for the unselected training samples to preserve information. In our method, we assign different weights to different training samples, in which higher weight indicates being selected. Then, the weight is used as interpolation coefficient to interpolate pseudo label vector and soft label vector. 

In epoch $t\!-\!1$, when using dataset A for task B, we obtain the soft label vector of $\bm{I}^a_i$ as $\bm{\tilde y}_i^{b} = p_{t-2}^b(\bm{I}^a_i)$. We can easily obtain its pseudo label vector $\bm{\bar y}_i^{b}$  based on $\bar {y}_{i,k}^{b}=1$ if $k=\arg\max_{k'} \tilde{y}_{i,k'}^{b}$ and $\bar {y}_{i,k}^{b}=0$ otherwise, in which $\tilde{y}_{i,k}^{b}$ (\emph{resp.}, $\bar{y}_{i,k}^{b}$) is the $k$-th entry of  $\bm{\tilde y}_i^{b}$ (\emph{resp.}, $\bm{\bar y}_i^{b}$). Assume we have learnt the weight $w_i$ for $\bm{I}^a_i$, the interpolated label vector is

\vspace{-8pt}
\begin{equation}
\begin{aligned}
\bm{\hat y}^b_i= w_i \cdot \mathbf{\overline y}^{b}_i + (1-w_i) \cdot \mathbf{\tilde y}^{b}_i. \label{interpolation_b}
\end{aligned}
\end{equation}
Then, in epoch $t\!-\!1$, the loss function can be written as

\vspace{-15pt}
\begin{equation}
\begin{aligned}
\min \limits_{\bm{\theta}^s,\bm{\theta}^a,\bm{\theta}^b} \sum_{i=1}^{n_a} L(\bm{y}_i^{a},p_{t-1}^a(\bm{I}^a_i)) + L(\bm{\hat y}_i^b,p_{t-1}^b(\bm{I}^a_i)). \label{proposed_loss_a}
\end{aligned}
\end{equation}

By comparing (\ref{proposed_loss_a}) with (\ref{naive_loss_a}), the only difference is that $\bm{\tilde y}_i^b$ is replaced with $\bm{\hat y}_i^b$. So (\ref{naive_loss_a}) is a special case of (\ref{proposed_loss_a}) when $w_i=0$. 
Similarly, in epoch $t$, the loss function of our method is

\vspace{-15pt}
\begin{equation}
\begin{aligned}
\min \limits_{\bm{\theta}^s,\bm{\theta}^a,\bm{\theta}^b} \sum_{i=1}^{n_b} L(\bm{\hat y}^{a}_i,p_{t}^a(\bm{I}^b_i)) + L(\bm{y}^b_i,p_{t}^b(\bm{I}^b_i)), \label{proposed_loss_b}
\end{aligned}
\end{equation}
in which $\bm{\hat y}^{a}_i$ can be obtained similar to $\bm{\hat y}^b_i$ in (\ref{interpolation_b}).

%Therefore, for each task, we tend to use training samples in the unlabeled dataset with pseudo labels to add information instead of preserving information. Specifically, 
Before each epoch, we learn weight $w_i$ for each training sample, which is used to interpolate its pseudo label vector and soft label vector. This is equivalent to using $w_i$ to control the tradeoff between adding and preserving information. The remaining problem is how to determine $w_i$, which will be detailed in the following sections. 
%-------------------------------------------------------
\subsection{Data Selection} \label{sec:data_selection}
Since alternating training strategy is adopted, we take training epoch $t$ using dataset B as an example to describe our method in Section 4.1 and 4.2. As depicted in Figure~\ref{fig:training_strategy}, in epoch $t$, training images $\mathcal{D}_B=\{\mathbf{I}^b_1, ..., \mathbf{I}^b_{n_b}\}$ from dataset B with labels $\mathcal{Y}_B=\{\mathbf{y}^b_1, ..., \mathbf{y}^b_{n_b}\}$ are fed into the multi-task network. To augment task A by using dataset B with pseudo labels, we tend to select the training samples in dataset B with confident pseudo labels and close data distribution to dataset A. 
%On one hand, we select the training samples in dataset B with confident pseudo labels based on the decision values in soft label vector and the local density of training samples. On the other hand,  we select the training samples in dataset B with close data distribution to dataset A based on the data distribution difference between dataset A and dataset B.

%--------------------------
\subsubsection{Selecting Data with Confident Pseudo Labels} \label{sec:sel_con}
Given an image $\mathbf{I}^b_i$ with its pseudo label vector $\bm{\bar y}_i^a$ generated from its soft label vector $\bm{\tilde y}_i^a$, one intuitive measurement of the confidence of $\bm{\bar y}_i^a$ is the highest decision value in $\bm{\tilde y}_i^a$, which reflects the probability of $\mathbf{I}_i^b$ being classified into its pseudo category $\arg\max_k \tilde{y}_{i,k}^a$. Formally, the confidence of pseudo label vector $\bm{\tilde y}_i^a$ can be measured by $
w^c_i = \max \limits_{k} \tilde {y}_{i,k}^a$.

However, $w^c_i$ does not take the relation among different training samples into account and thus its reliability is significantly compromised. Inspired by recent work~\cite{guo2018curriculumnet} which leverages local density to measure the purity of noisy labels, we also assume that the training images with higher local density are more prone to have correct pseudo labels. Due to space limitation, we leave the details of calculating local density to Supplementary. Finally, we combine $w^c_i$ with normalized local density $w^d_i$  to measure the confidence of pseudo label: $w^s_i = w^c_i \cdot w^d_i$.
%--------------------------

\vspace{-2pt}
\subsubsection{Selecting Data with Closer Data Distribution} \label{sec:sel_dis}
In order to bridge the domain gap between dataset A and dataset B, we can select samples from dataset B with closer data distribution to dataset A to facilitate task A. In this paper, as a simple approach, we group the training samples in dataset B into $C^b$ clusters and find the clusters with closer data distribution to dataset A. Considering the data variance within dataset A, it may be ineffective to calculate the domain difference between each cluster and the entire dataset A. So we also group the training samples in dataset A into $C^a$ clusters, and calculate the domain difference between each cluster in dataset B and each cluster in dataset A based on Maximum Mean Discrepancy (MMD)~\cite{Huang2007KMM}, which is commonly used in domain adaptation~\cite{PatelGLC15}. Then, we calculate the weighted sum of distances between each cluster from dataset B and the entire dataset A, in which the weights can be learnt by Earth Mover's Distance (EMD). 
%Finally, different weights $w^g$ are assigned to different samples in dataset B based on the difference of data distribution, which can be measured by distance between dataset A and dataset B. 

Due to space limitation, we leave the details of calculating EMD to Supplementary. We use $d_k^{E}$ to denote EMD between the $k$-th cluster in domain B and the whole domain A, and use $\gamma_{i,k}$ to denote the probability that $i$-th sample is from the $k$-th cluster. Then, the distance between each sample in  domain B and the entire  domain A can be calculated as $\hat{d}_i = \sum_{k=1}^{C^b} d_k^{E} \cdot \gamma_{i,k}$.
Finally, we tend to assign weight $w^g_i =\exp(-\lambda \cdot \hat{d}_i)$ for the $i$-th sample in domain B to select those close to domain A, where $\lambda$ is set as $0.1$ in all experiments.

%--------------------------
\subsection{Label Vector Interpolation} \label{sec:label_vector_interpolation}
%As mentioned before, when using dataset B to augment task A in epoch $t$, pseudo label vector can be used to add information while soft label vector can be used to preserve information. 
We aim to select the training samples in dataset B with confident pseudo labels and close data distribution to dataset A to add information, by assigning larger weights on their pseudo label vectors. Based on previously introduced $w^s_i$ and $w^g_i$, the combined weight can be calculated and normalized by $w_i = \frac {w^s_i \cdot w^g_i} {\max \limits_{j\in[1,n_b]} w^s_j \cdot w^g_j}$.
Then, we can arrive at the interpolated label vector:

\vspace{-10pt}
\begin{equation}
\begin{aligned}
\bm{\hat y}^a_i= w_i \cdot \mathbf{\overline y}^{a}_i + (1-w_i) \cdot \mathbf{\tilde y}^{a}_i, \label{interpolation_a}
\end{aligned}
\end{equation}
which corresponds to the interpolated label vector in (\ref{proposed_loss_b}). 
Inspired by knowledge distillation~\cite{Hinton2015Distilling}, we make slight improvement on the soft label vector $\mathbf{\tilde y}^{a}_i$ by introducing the factor of temperature $T$, and arrive at $\bm{\tilde y}^{a'}_i$ with the $j$-th element being $\tilde{y}^{a'}_{i,j}= \frac {(\tilde{y}^{a}_{i,j})^{1/T}} {\sum_{j=1}^{C^a} (\tilde{y}^{a}_{i,j})^{1/T}}$,
in which $\tilde{y}^{a}_{i,j}$ is the $j$-th element in $\mathbf{\tilde y}^{a}_i$.
%As suggested in~\cite{Hinton2015Distilling}, setting the temperature $T > 1$ can increase the weight of smaller logit values and encourage the network to better encode similarities among categories, which is also confirmed in~\cite{li2018learning}. 
In our experiments, we replace $\mathbf{\tilde y}^{a}_i$ in (\ref{interpolation_a}) by $\bm{\tilde y}^{a'}_i$ with $T=2$, which can generally achieve good performance. The summary of whole training algorithm is left to Supplementary due to space limitation.

\section{Experiment}

\setlength{\tabcolsep}{10pt}
\begin{table*}[t]
\small
  \caption{Accuracy(\%) of different methods on four pairs of datasets. Best results are denoted in boldface.}\smallskip 
  \centering
  %\fontsize{8}{8}\selectfont
  \begin{tabular}{|c|c|c||c|c||c|c||c|c|}
      \hline
      \multirow{2}{*}{Method}&
      \multicolumn{6}{c|}{Face-centric}&\multicolumn{2}{c|}{Human-centric}\cr
      \cline{2-9}&Expw &AFLW &FER+ &AFLW &SFEW &AFLW &DeepFashion &PETA \cr
  \hline\hline
    STL&64.17&75.17&82.13&75.17&47.49&75.17&81.71&78.08\cr
    \hline\hline
    % All-in-one~\cite{ranjan2017all}& 63.54&76.01 & 82.57&76.41&45.51 &75.27&80.39& 79.02\cr
    All-in-one& 63.54&76.01 & 82.57&76.41&45.51 &75.27&80.39& 79.02\cr
    \hline
    % LwF~\cite{li2018learning}& 66.10&77.08& 82.73&76.12 &50.11 &76.02&82.12&79.83\cr
    MTL-wF& 66.10&77.08& 82.73&76.12 &50.11 &76.02&82.12&79.83\cr
    \hline
    \hline
    % Logits~\cite{misra2016seeing}&0.6598 & 0.7689& 0.8298&0.7681&0.5031 &0.7641\cr
    % SFSMR~\cite{chang2017semisupervised}& 65.04&76.81 & 82.89&76.78&51.01 &76.07&81.83&80.04 \cr
    SFSMR& 65.04&76.81 & 82.89&76.78&51.01 &76.07&81.83&80.04 \cr
    \hline
    % SLRM~\cite{jing2015semi}& 65.44&76.51 & 82.91&76.82&50.61 &75.92&82.01&79.43 \cr
    SLRM& 65.44&76.51 & 82.91&76.82&50.61 &75.92&82.01&79.43 \cr
    % SFSMR~\cite{lu2015semi}& & & & &  &\cr
    \hline
    % LEL-LTN~\cite{augenstein2018multi}&65.42 &76.17 &83.08 &75.89 & 50.56 &76.49&81.23&79.84 \cr
    LEL-LTN&65.42 &76.17 &83.08 &75.89 & 50.56 &76.49&81.23&79.84 \cr
    \hline
    % DCN-AP~\cite{zhang2018facial}&66.01 & 77.17&82.14 & 76.75& 52.10 &76.54&82.93&80.11\cr
    DCN-AP&66.01 & 77.17&82.14 & 76.75& 52.10 &76.54&82.93&80.11\cr
    \hline\hline
    % SSMTC~\cite{wang2009semi}& & & & &  &\cr
    % Liu etal~\cite{liu2008semi}& & & & &  &\cr
    MTL-SA&$\mathbf{67.34}$ &$\bm{78.41}$ &$\bm{84.45}$ & $\bm{77.92}$&$\bm{53.50}$ &$\bm{77.85}$ &$\mathbf{84.12}$ &$\bm{81.78}$ \cr
    \hline
  \end{tabular}
%   \vspace{0.1mm}
  \label{tab:performance_comparison}
  \vspace{-10pt}
\end{table*}
%.
%-------------------------------------------------------------------------

\subsection{Datasets}
%Existing MTC methods have been applied to a wide range of tasks~\cite{zhang2014facial,han2018disjoint,ranjan2019hyperface,lu2017fully,sarafianos2017curriculum}. In this paper, we focus on face-centric tasksincluding facial expression recognition and pose estimation, and human-centric tasks including cloth style classification and human attribute classification. As 
For face-centric applications, we construct three pairs of disjoint datasets by using three facial expression datasets of different scales (\emph{i.e.}, Expw~\cite{zhang2018facial}, FER+~\cite{barsoum2016training}, and SFEW~\cite{dhall2011static}), and one pose dataset (\emph{i.e.}, AFLW~\cite{koestinger2011annotated}) for facial expression recognition and pose estimation. We also construct one pair of disjoint human-centric datasets by using one clothes style dataset (\emph{i.e.}, DeepFashion~\cite{liuLQWTcvpr16DeepFashion}) and one human attribute dataset (\emph{i.e.}, PETA~\cite{deng2014pedestrian}) for cloth style classification and age stage estimation. The details of datasets and training/test splits can be found in Supplementary.

\subsection{Implementation Details}
Following~\cite{pons2018multi}, we use the convolutional layers of VGG as shared layers, and two FC layers as task-specific layers for two tasks. We also employ cross-stitch layer~\cite{misra2016cross} between FC layers of two tasks. For fair comparison, we use the same backbone network for all methods.
%, which can highlight the performance gain attributed to different optimization methods rather than different network configurations. 
In the training stage, we set the batchsize as $32$ and use Adam optimizer with the learning rate  $0.0001$.

\subsection{Comparison with Other Multi-task Learning for Classification Methods}
In this section, we compare the performance of MTL-SA with three groups of baselines. In the first group, we compare with All-in-one network~\cite{ranjan2017all} using joint training strategy and MTL-wF using alternating training strategy. In the second group, we compare with manifold based semi-supervised MTL methods, including SFSMR~\cite{chang2017semisupervised} and SLRM~\cite{jing2015semi}. In the third group, we compare with semi-supervised MTL methods LEL-LTN~\cite{augenstein2018multi} and DCN-AP~\cite{zhang2018facial} using pseudo labels. We also compare with Single-Task Learning (STL), which uses one separate network for each task without parameter sharing. The details of baselines are provided in Supplementary.

The results of different methods are summarized in Table~\ref{tab:performance_comparison}. Based on Table~\ref{tab:performance_comparison}, we observe that MTL methods generally outperform STL, which demonstrates the benefit of sharing information among multiple tasks. We also observe that LwF~\cite{li2018learning} with alternating training strategy achieves better results than All-in-one~\cite{ranjan2017all} with joint training strategy, which indicates the advantage of alternating training strategy. Another observation is that semi-supervised MTL methods SFSMR~\cite{chang2017semisupervised}, SLRM~\cite{jing2015semi}, LEL-LTN~\cite{augenstein2018multi} and DCN-AP~\cite{zhang2018facial} generally outperform All-in-one~\cite{ranjan2017all}, which shows that it is helpful to use unlabeled training samples based on manifold structure information or refined pseudo label information. 
It can also be seen that our proposed MTL-SA achieves significant improvement over the closest related baseline MTL-wF, which demonstrates the advantage of selectively adding information instead of merely preserving information. Moreover, our method achieves the best results on all four pairs of datasets, which indicates the effectiveness of selectively augmenting each task  by using the training samples in the unlabeled dataset with confident pseudo labels and close data distribution to the labeled dataset.

\subsection{Ablation Studies}

%We also include the results of MTL-wF from Table~\ref{tab:performance_comparison} because MTL-wF is closely related to MTL-SA ($w^i=0$) and the only difference is that  MTL-SA ($w^i=0$) uses knowledge distillation with temperature $T=2$. Finally, we include the results of our method from Table~\ref{tab:performance_comparison} for comparison.

\noindent
\setlength{\tabcolsep}{4pt}
\begin{table}[t]
\small
  \caption{Accuracy(\%) of our special cases on two pairs of datasets. Best results are denoted in boldface.}

  \centering
  %\fontsize{8}{8}\selectfont
    \begin{tabular}{|c|c|c|c|c|}
      \hline
      \multirow{2}{*}{Method}&
      \multicolumn{2}{c|}{Face-centric} & \multicolumn{2}{c|}{Human-centric}\cr\cline{2-5}
      &FER+ &AFLW &DeepFashion &PETA \cr
    \hline
    \hline
    % LwF~\cite{li2018learning}& 82.73&76.12 &82.12&79.83 \cr
    MTL-wF& 82.73&76.12 &82.12&79.83 \cr
    \hline
    \hline
    $w=0$ & 82.88&76.23  &82.53&80.04  \cr
    \hline
    $w=1$ &82.48 &75.62  &81.12 &79.14  \cr
    \hline
    $w=0.5$&82.74& 76.17 &81.79 &79.46\cr
    \hline
    only $w^c$&83.01 &76.92  & 82.45& 79.98\cr
    \hline
    only $w^d$&83.21 &77.01  &82.71 & 79.96\cr
    \hline
    \hline
    only $w^g$(EMD) &83.44 &77.42 &82.94 & 80.32\cr
    \hline
    only $w^g$(MMD) &83.02 &77.12 &82.46 &80.03 \cr
    \hline
    \hline
    MTL-SA&$\bm{84.45}$ &$\bm{77.92}$ &$\bm{84.12}$ &$\bm{81.78}$ \cr
    \hline
  \end{tabular}
%   \vspace{0.1mm}
  \label{tab:component analysis}
  \vspace{-20pt}
\end{table}
Note that the final weight used in our method $w_i=w_i^c\cdot w_i^d\cdot w_i^g$ is based on three types of weights $w_i^c$, $w_i^d$, and $w_i^g$. To investigate the importance of each type of weight, we perform ablation studies on our MTL-SA method. By taking a pair of face-centric datasets and a pair of human-centric datasets as examples, we report the results of three special cases with constant weights by setting $w^i$ as $0$, $1$, and $0.5$. When $w^i=0$, we only use knowledge distillation to preserve the information obtained in the previous epoch. When $w^i=1$, we use pseudo labels for all training samples. When $w^i=0.5$, we use simple label interpolation with the same weight for all training samples. Besides, we report the results of three special cases only using one type of weight (\emph{i.e.}, $w_i^c$, $w_i^d$, or $w_i^g$).
Experimental results are summarized in Table~\ref{tab:component analysis}. We observe that  MTL-SA ($w^i=0$) is slightly better than MTL-wF~\cite{li2018learning}, which indicates the benefit of knowledge distillation with higher temperature $T>1$. We also observe that MTL-SA ($w^i=1$) achieves worse results compared with MTL-SA ($w^i=0$), which shows that it is harmful to use all training samples with pseudo labels due to the label noise and data distribution mismatch. By comparing our special cases only using one type of weight (\emph{i.e.}, MTL-SA (only $w^c$), MTL-SA (only $w^d$),  MTL-SA (only $w^g$)) with the special cases using constant weights (\emph{i.e.}, MTL-SA ($w=0$), MTL-SA ($w=1$), MTL-SA ($w=0.5$)), it can be seen that simple interpolation of pseudo label vector and soft label vector with a constant weight is not very effective while our special cases using any type of weight generally outperform the simple interpolation. Among three types of weights, MTL-SA (only $w^g$) performs more favorably, which might be because that the domain gap between FER+ (\emph{resp.}, DeepFashion) and AFLW (\emph{resp.}, PETA) is quite huge and can be mitigated by selecting the training samples with close data distribution. Finally, our full-fledged MTL-SA method outperforms all special cases on both pairs of datasets, which verifies the effectiveness of selecting training sample in the unlabeled dataset to augment each task based on multiple selection criteria. We have similar observations on the other pairs of datasets. 

Recall that we use EMD to measure distribution difference. We also compare with MTL-SA (only $w^g$(MMD)) which simply uses MMD instead of EMD. We observe that MTL-SA (only $w^g$(MMD)) underperforms MTL-SA (only $w^g$(EMD)), which indicates the benefit of our design.

\subsection{Qualitative Results}
In Supplementary, we provide in-depth analyses to prove the existence of two issues: pseudo label noise and data distribution mismatch. We also show qualitative results to corroborate the effectiveness of our method to address the above two issues.

\section{Conclusion}
In this paper, we have proposed our  Multi-Task Learning for Classification with Selective Augmentation (MTL-SA) method  based on our designed three sample selection criteria.
%to select the training samples in the unlabeled dataset with confident pseudo labels and close data distribution to the labeled dataset. 
Comprehensive experiments on three pairs of face-centric datasets and one pair of human-centric datasets have demonstrated the effectiveness of our proposed method.

\renewcommand\thesection{\Alph{section}}

In this Supplementary, we first describe the technical details that are omitted in the main text in Section~\ref{sec:local_density}, \ref{sec:EMD}, \ref{sec:train_alg}. Then, we introduce the details of our used datasets and compared baselines in Section~\ref{sec:datasets}, \ref{sec:baselines}. Finally, we provide more experimental results in Section~\ref{sec:in-depth}, \ref{sec:qualitative}.

\section{Details of Calculating Local Density}~\label{sec:local_density}
In Section 4.1.1 in the main text, we proposed to use local density as the measurement of pseudo label confidence and here we provide the technical details. We group images $\mathcal{D}_B=\{\mathbf{I}^b_1, ..., \mathbf{I}^b_{n_b}\}$ in dataset B  based on their pseudo labels. For the $c$-th group of training samples with the same pseudo label $c$, we use feature extractor $f_{t-1}^a(\cdot)$ (\emph{i.e.}, model parameters $\{\bm{\theta}^{s},\bm{\theta}^{a}\}$ for task A with the last classification layer removed) to extract their features. Note that in this section, we use the feature extractor corresponding to the task (\emph{i.e.}, task A) which needs to be augmented with the current unlabeled dataset (\emph{i.e.}, dataset B). Then, we calculate the Euclidean distance matrix $ \bm{D} \in \mathcal{R}^{n^b_c \times n^b_c}$ with $n^b_c$ being the number of samples in the $c$-th group. Each entry in $\bm{D}$ is calculated by $D_{i,j} = || f_{t-1}^a(\bm{I}^b_i)- f_{t-1}^a(\bm{I}^b_j) ||^2$.
Given an image $\mathbf{I}^b_i$ in the $c$-th group, we calculate its local density as $\rho_i= \sum_{j} \delta(D_{i,j}<d_c)$,
where $\delta(D_{i,j}<d_c)=1$ if $D_{i,j}<d_c$ and $0$ otherwise, in which $d_c$ is determined by sorting $(n^b_c)^2$ entries in $ \bm{D}$ in increasing order and select the number at the location $\lceil \kappa\cdot (n^b_c)^2 \rceil$ ($\kappa=0.6$ in our experiments). Hence, $\rho_i$ is the number of samples in the $c$-group with the distance to  $\mathbf{I}^b_i$ smaller than $d_c$. We assume the images in each group with correct pseudo labels should be close to each other in the feature space, leading to a large value of local density. Therefore, for $\mathbf{I}^b_i$ in the $c$-th group, we use normalized local density $w^d_i = \frac{\rho_i}{\max \limits_{j\in \mathcal{I}_c}\rho_{j}}$  to measure its pseudo label confidence, in which $\mathcal{I}_c$ is the index set of the $c$-th group. 

\section{Details of Calculating Earth Moving Distance (EMD) between Two Domains}\label{sec:EMD}
In Section 4.1.2 in the main text, we proposed to calculate the data distribution difference between each cluster in dataset B and the entire dataset A. Formally, we treat the training samples $\mathcal{D}_B=\{\mathbf{I}^b_1, ..., \mathbf{I}^b_{n_b}\}$ (\emph{resp.},  $\mathcal{D}_A=\{\mathbf{I}^a_1, ..., \mathbf{I}^a_{n_a}\}$) in dataset B (\emph{resp.}, A) as domain B (\emph{resp.}, A). We use feature extractor $f^a_{t-1}(\cdot)$ to extract features for $\mathcal{D}_B$ and $\mathcal{D}_A$. Based on the extracted features, we use Gaussian Mixed Model (GMM) to group domain B (\emph{resp.}, A) into $C^b$ (\emph{resp.}, $C^a$) clusters, in which each cluster is a Gaussian model with the mean feature vector $\bm{\mu}^b_k$ (\emph{resp.}, $\bm{\mu}^a_k$) and prior weight $\bm{\pi}^b_k$ (\emph{resp.}, $\bm{\pi}^a_k$). Thus, we can define domain B (\emph{resp.}, A) as  $\{(\bm{\mu}^b_1,\bm{\pi}^b_1),...,(\bm{\mu}^b_{C^b},\bm{\pi}^b_{C^b})\}$ (\emph{resp.}, $\{(\bm{\mu}^a_1,\bm{\pi}^a_1),...,(\bm{\mu}^a_{C^a},\bm{\pi}^a_{C^a})\}$). We calculate the MMD distance $d^M_{k,j}$ between the $k$-th cluster in domain B and the $j$-th cluster in domain A as $d_{k,j}^M =|| {\bm{\mu}^b_k} - {\bm{\mu}^a_j} ||^2$. Furthermore, we can obtain the distance between the $k$-th cluster in  domain B and the entire domain A, which is calculated by Earth Mover's Distance (EMD) between $\{(\bm{\mu}^b_k,1)\}$ and $\{(\bm{\mu}^a_1,\bm{\pi}^a_1),...,(\bm{\mu}^a_{C^a},\bm{\pi}^a_{C^a})\}$:
\begin{equation}
\begin{aligned}
d_k^{E}={\frac {\sum_{j=1}^{C^a}h_{k,j}d_{k,j}^M}{\sum_{j=1}^{C^a}h_{k,j}}},  \label{emd}
\end{aligned}
\end{equation}
where the optimal flow $h_{k,j}$ corresponding to the least amount of total work is obtained by solving the EMD optimization problem. The distance $d_k^{E}$ can be viewed as the required amount of work to move the $k$-th cluster from  domain B to the whole domain A.

\section{The Whole Training Algorithm}\label{sec:train_alg}
We rewrite the loss functions of our method as follows. The loss function in epoch $t\!-\!1$ can be written as
\begin{equation}
\begin{aligned}
\min \limits_{\bm{\theta}^s,\bm{\theta}^a,\bm{\theta}^b} \sum_{i=1}^{n_a} L(\bm{y}_i^{a},p_{t-1}^a(\bm{I}^a_i)) + L(\bm{\hat y}_i^b,p_{t-1}^b(\bm{I}^a_i)). \label{proposed_loss_a}
\end{aligned}
\end{equation}
Besides, the loss function in epoch $t$ can be written as
\begin{equation}
\begin{aligned}
\min \limits_{\bm{\theta}^s,\bm{\theta}^a,\bm{\theta}^b} \sum_{i=1}^{n_b} L(\bm{\hat y}^{a}_i,p_{t}^a(\bm{I}^b_i)) + L(\bm{y}^b_i,p_{t}^b(\bm{I}^b_i)). \label{proposed_loss_b}
\end{aligned}
\end{equation}

The whole training algorithm is summarized in Algorithm~\ref{alg::algorithm}. In the testing stage, we use $\{\bm{\theta}^s, \bm{\theta}^a\}$ (
\emph{resp.}, $\{\bm{\theta}^s, \bm{\theta}^b\}$) for  evaluation on the task A (\emph{resp.}, B).

 \begin{algorithm}[t]
     \caption{Alternating Optimization Algorithm for MTL-SA}
     \label{alg::algorithm}
     \begin{algorithmic}[1] 
         \Require Training images $\mathcal{D}_a$ with labels $\mathcal{Y}_a$ in dataset A and training images $\mathcal{D}_b$ with labels $\mathcal{Y}_b$ in dataset B. Model parameters $\bm{\theta}^s$, $\bm{\theta}^a$, and $\bm{\theta}^b$ initialized by joint training method~\cite{ranjan2017all}.
         \For{$t = 1 \to t_{max}$}
         \If{$t \% 2=0$}
            \State Feed $\mathcal{D}_b$ into the network to obtain $f_{t-1}^a(\mathbf{I}^b)$, $\bm{\tilde y}^{a}$, and $\bm{\bar y}^{a}$.
            \State Calculate $w^s$ for $\mathcal{D}_b$.
            \State Calculate $w^g$ for $\mathcal{D}_b$.
            \State Calculate $w$ for $\mathcal{D}_b$.
            \State Calculate the interpolated label vector $\hat{\bm{y}}^a$.
            \State Update $\bm{\theta}^s$, $\bm{\theta}^a$, and $\bm{\theta}^b$ based on the loss function in Eqn. (\ref{proposed_loss_b}).
        \Else
            \State Execute Line 3-7 with $a$ and $b$ exchanged.
            \State Update $\bm{\theta}^s$, $\bm{\theta}^a$, and $\bm{\theta}^b$ based on the loss function in Eqn. (\ref{proposed_loss_a}).
        \EndIf
       \EndFor
       \State \Return{Model parameters $\bm{\theta}^s$, $\bm{\theta}^a$ and $\bm{\theta}^b$}.
     \end{algorithmic}
 \end{algorithm}

\section{Details of Datasets} \label{sec:datasets}
For face-centric tasks, we use three datasets for facial expression recognition with different scales including a large-scale dataset Expw~\cite{zhang2018facial}, a medium-scale dataset FER+~\cite{barsoum2016training}, and a small-scale dataset SFEW~\cite{dhall2011static}. In detail, Expw has over 90,000 images collected from websites, which are split into $80\%$ training images and $20\%$ testing images.  FER+ contains 28,709 training images, 3,589 validation images, and 3,589 testing images. SFEW consists of a training set with 958 samples, a validation set with 436 samples, and a test set with 372 samples. Expw and SFEW are annotated with seven expression labels while FER+ is labeled with eight expression labels. Besides, the facial poses in these datasets are quite diverse, so we conjecture that the facial pose information from pose datasets could be beneficial for the facial expression recognition task. For facial pose estimation, we use AFLW dataset~\cite{koestinger2011annotated}, which totally has 25,993 faces labeled with five types of poses including left profile face, left face, frontal face, right face and right profile face. We divide AFLW into $80\%$ training samples and $20\%$ test samples following~\cite{zavan2018benchmarking}. Based on the above mentioned three facial expression datasets (\emph{i.e.}, Expw, FER+, and SFEW) and one facial pose dataset (\emph{i.e.}, AFLW), we construct three pairs of facial expression and pose datasets, leading to three MTL settings with two disjoint datasets for two tasks.

For human-centric tasks, we construct one pair of disjoint datasets for clothes style classification and human attribute (\emph{e.g.}, age stage) estimation. Our used dataset for clothes style classification is DeepFashion~\cite{liuLQWTcvpr16DeepFashion} with 289, 222 images from 50 clothes styles, and we splite it into training samples, validation samples and tesing samples according to the ratio of $4:1:1$. Our used dataset for age stage estimation is PETA~\cite{deng2014pedestrian} with 19, 000 images from 4 age stages (16 to 30, 31-45, 46-60, and above 61), and the split rule is the same as above used in DeepFashion dataset.

\section{Details of Baselines}\label{sec:baselines}
We compare our MTL-SA method with three groups of baselines. 

For the first group of baselines, we compare with All-in-one network~\cite{ranjan2017all} using joint training strategy and MTL-wF using alternating training strategy. In particular, to compare with All-in-one~\cite{ranjan2017all}, we mix training samples from two datasets to train our model in each epoch. To compare with MTL-wF, we use one dataset to train our network in each epoch and alternate between two datasets, in which soft label vectors are used to prevent forgetting effect. 

For the second group of baselines, we compare with manifold learning based semi-supervised MTL methods ~\cite{lu2015semi,chang2017semisupervised,jing2015semi}, which utilizes manifold regularization on unlabeled training samples. Specifically, to compare with SFSMR~\cite{chang2017semisupervised}, we use $l_{2,1}$-norm and trace norm term to generate manifold regularization for label fitness and manifold smoothness. To compare with SLRM~\cite{jing2015semi}, we combine nuclear norm and Laplacian norm for complexity regularization and smoothness regularization. 

For the third group of baselines, we compare with semi-supervised MTL methods LEL-LTN~\cite{augenstein2018multi} and DCN-AP~\cite{zhang2018facial}, which use pseudo labels to boost multiple tasks. 
Particularly, to compare with LEL-LTN~\cite{augenstein2018multi} which employs label transfer network to tag samples with pseudo labels, we add a label transfer module to the penultimate layer of our network. To compare with DCN-AP~\cite{zhang2018facial} which uses label propagation to fill in missing labels similar to multi-label learning, we leverage Markov Random Field (MRF) to refine the pseudo labels based on our predicted labels. Finally, we also compare with Single-Task Learning (STL), which uses one separate network for each task without parameter sharing.

% \subsection{Ablation Studies}

% To investigate the importance of each type of weight further, we also perform ablation studies of our SA-MTL method on DeepFashion and PETA datasets for human-centric tasks, we report the results of three special cases with constant weights by setting $w^i$ in (\ref{total_weights}) as $0$, $1$, and $0.5$. When $w^i=0$. By comparing such three special cases, it can be seen that simple interpolation of pseudo label vector and soft label vector with a constant weight is not very effective while our special cases using any type of weight generally outperform the simple interpolation, which indicates that it is useful to select training samples for task augmentation based on the decision value, local density, or data distribution distance. Among three types of weights, SA-MTL (only $w^g$) performs more favorably, which might be because that the domain gap between DeepFashion and PETA is quite huge and can be mitigated by selecting the training samples with close data distribution.

% Besides, we also compare MMD and EMD distribution measurement in special case SA-MTL(only $w^g$), we can observe that the result of SA-MTL(only $w^g$(MMD)) is worse than our result using EMD. 

\begin{figure*}[h]
\begin{center}
\includegraphics[scale=0.48]{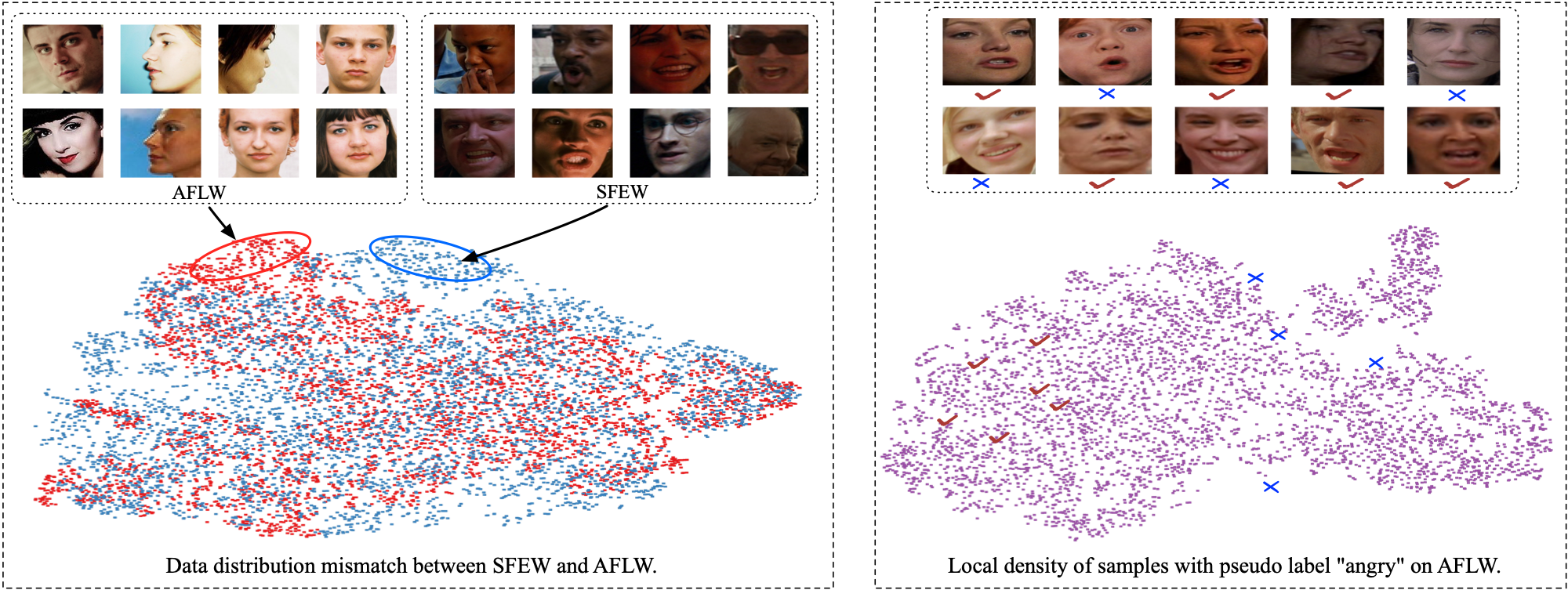}
\end{center}
\caption{Given one task (\emph{i.e.}, facial expression recognition) with a labeled dataset (\emph{i.e.}, SFEW) and an unlabeled dataset (\emph{i.e.}, AFLW), there are two issues when using AFLW with pseudo labels to augment this task: 1) data distribution mismatch between AFLW and SFEW; 2) noisy pseudo labels of AFLW dataset. Image samples are visualized with t-SNE based on their features. Best viewed in color.}
\label{fig:flowchat}
\end{figure*}

\section{In-depth Analyses of Data Distribution Mismatch and Local Density}\label{sec:in-depth}

We visualize sampled images from AFLW and SFEW datasets with t-SNE based on their extracted features in Figure~\ref{fig:flowchat}. In the left subfigure, the data distributions of two datasets are overlapped to certain extent but still considerably different. Visually, the images in SFEW are often captured in poor light condition while the images in dataset AFLW are generally captured in good light condition, resulting in the data distribution discrepancy between these two datasets.

SFEW is a facial expression dataset while AFLW is a pose dataset, so AFLW does not have ground-truth emotion labels. But we can get pseudo emotion labels for images in AFLW through our method. In the right subfigure, we show a few images from AFLW with the pseudo emotion label ``angry", in which some images are actually not ``angry". This indicates the existence of pseudo label noise. However, based on the right subfigure, the samples with correct pseudo labels are more prone to have higher local density.

% \vspace{-10pt}
% \begin{table}[h]  
%   \caption{Accuracy(\%) of different methods on DeepFashion and PETA datasets. Best results are denoted in boldface.}
%     \centering
%     \fontsize{8}{8}\selectfont
%         \begin{tabular}{|c|c|c|}
%             \hline
%             \multirow{2}{*}{Method}&
%             \multicolumn{2}{c|}{Dataset}\cr\cline{2-3}
%             &DeepFashion &PETA \cr
%         \hline
%     \hline
%     DML-LWF&82.12&79.83 \cr
%     \hline
%     \hline
%     SA-MTL ($w=0$) &82.53&80.04  \cr
%     \hline
%     SA-MTL ($w=1$) &81.12 &79.14  \cr
%     \hline
%     SA-MTL ($w=0.5$)&81.79 &79.46\cr
%     \hline
%     SA-MTL (only $w^c$  &82.04 & 79.87\cr
%     \hline
%     SA-MTL (only $w^d$ &82.18 &79.85 \cr
%     \hline
%     SA-MTL (only $w^g$(EMD))  &82.94 & 80.32\cr
%     \hline
%     SA-MTL (only $w^g$(MMD)) &81.98 &79.85 \cr
%     \hline
%     \hline
%     SA-MTL&$\mathbf{84.12}$ &$\bm{81.78}$ \cr
%     \hline
%     \end{tabular}
%     \label{tab:component analysis}
%     \vspace{-12pt}
% \end{table}

Next, we attempt to investigate data distribution mismatch and pseudo label confidence in a quantitative way.
For data distribution mismatch, we calculate MMD~\cite{Huang2007KMM} between AFLW and FER+ datasets, and the MMD value is $0.243$. We also divide AFLW into two clusters using K-means and calculate MMD between two clusters. The obtained MMD value is $0.031$, which is much lower than $0.243$, which proves the existence of data distribution mismatch between two datasets. 

One measurement of pseudo label confidence is local density. We assume that the samples with high local density are more prone to have correct pseudo labels. From the images with pseudo ``fear" label on AFLW, we randomly sample $100$ images with local density larger than $0.95$ and $100$ images with local density smaller than $0.15$. Because AFLW does not provide ground-truth emotion labels, we manually annotate the emotion labels of selected samples to measure the accuracy of pseudo labels. We find that $96$ (\emph{resp.}, $3$) out of $100$ samples with high (\emph{resp.}, low) local density have correct pseudo labels, which proves the correlation between low local density and noisy pseudo labels. 

\begin{figure}[t]
\begin{center}
\includegraphics[scale=0.4]{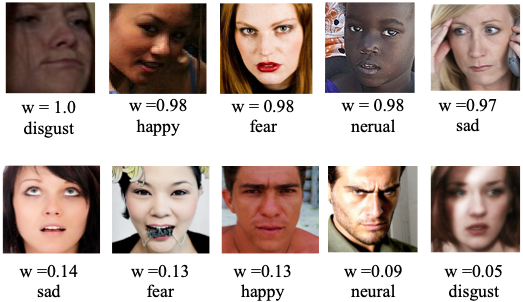}
\end{center}
\caption{Illustration of training samples in AFLW dataset with weights $w_i$ and pseudo facial expression labels. In the top (\emph{resp.}, bottom) row, we show five training sample with the highest (\emph{resp.}, lowest) weights obtained by our method .}
\label{fig:results}
\end{figure}

\section{Qualitative Analyses of Sample Selection} \label{sec:qualitative}
To investigate the effectiveness of our sample selection method in a qualitative way, we take the pair of SFEW and AFLW datasets as an example and show five images in AFLW dataset with the highest (\emph{resp.}, lowest) weights $w_i$ obtained by our method in the top (\emph{resp.}, bottom) row in Figure~\ref{fig:results}. From Figure~\ref{fig:results}, we observe that the pseudo facial expression labels of images in the top row are all correct while those in the bottom row are generally incorrect, which shows that the combination of decision value and local density is a reliable measurement for the confidence of pseudo labels. We also observe that some images in the top row are captured in dark environment. Based on our observation, the images in SFEW are often captured in poor light condition while the images in dataset AFLW are generally captured in good light condition, so the training images in the top row are visually more similar to the SFEW dataset, which implies the effectiveness of sample selection based on data distribution distance.
\bibliographystyle{IEEEbib}
\bibliography{icme2020template}

\begin{small}
\bibliographystyle{IEEEbib}
\bibliography{icme2020template}
\end{small}

\end{document}